\documentclass[conference]{IEEEtran}
\IEEEoverridecommandlockouts
\usepackage{cite}
\usepackage{amsmath,amssymb,amsfonts}
\usepackage{algorithmic}
\usepackage{graphicx}
\usepackage{textcomp}
\usepackage{xcolor}
\usepackage{subcaption}
\usepackage{booktabs}
\usepackage{url}
\usepackage{amsmath}
\usepackage{amsfonts}
\usepackage{bm}

\def\BibTeX{{\rm B\kern-.05em{\sc i\kern-.025em b}\kern-.08em
    T\kern-.1667em\lower.7ex\hbox{E}\kern-.125emX}}
    
\usepackage{mathtools}

\begin{document}

\title{Symbolic Graph Intelligence: Hypervector Message Passing for Learning Graph-Level Patterns with Tsetlin Machines}

\author{\IEEEauthorblockN{Christian D. Blakely \thanks{Any statements, ideas, and/or
opinions expressed in this paper are strictly those of the author}}
\IEEEauthorblockA{\textit{Co-founder @ Bernoly AG} \\
\textit{Centre for AI Research @ University of Agder, Norway}\\
blakely@bernoly.com}}

\maketitle

\begin{abstract}
We propose a multilayered symbolic framework for general graph classification that leverages sparse binary hypervectors and Tsetlin Machines (TMs). Each graph is encoded through structured message passing, where node, edge, and attribute information are bound and bundled into a symbolic hypervector. This process preserves the hierarchical semantics of the graph through layered binding—from node attributes to edge relations to structural roles—resulting in a compact, discrete representation. We also formulate a local interpretability framework which lends itself to a key advantage of our approach being locally interpretable: predictions can be traced back to specific nodes and edges by decoding their influence in the bundled representation. We validate our method on TUDataset benchmarks, demonstrating competitive accuracy with strong symbolic transparency compared to neural graph models.

\end{abstract}

\begin{IEEEkeywords}
tsetlin machine, graph machine learning
\end{IEEEkeywords}

\section{Introduction}

Graph classification is a fundamental task in graph-based machine learning, where the goal is to assign a label or predict a target for an entire graph. This problem arises in a wide range of applications, from predicting molecular properties \cite{gilmer2017neural, xu2019powerful} and protein function \cite{borgwardt2005protein}, to analyzing social networks \cite{yanardag2015deep} and brain connectivity graphs \cite{ktena2018spectral}.

Mainstream approaches to graph classification typically rely on message-passing neural networks (MPNNs), including Graph Convolutional Networks (GCNs) \cite{kipf2016semi}, Graph Attention Networks (GATs) \cite{velickovic2018graph}, and variants such as the Graph Isomorphism Network (GIN) \cite{xu2019powerful}. These models operate by iteratively aggregating feature information from each node’s neighbors, then pooling node-level representations into a global graph-level embedding. Modern benchmark frameworks such as the TUDataset collection \cite{morris2020TUDataset} and Open Graph Benchmark (OGB) \cite{hu2020ogb} provide datasets where each sample is a graph with arbitrary topology and rich node/edge features. While effective, these approaches often involve complex architectures, are difficult to interpret, and can be sensitive to structural noise or training instability.

\begin{figure}  
    \centering
    \includegraphics[width=3in]{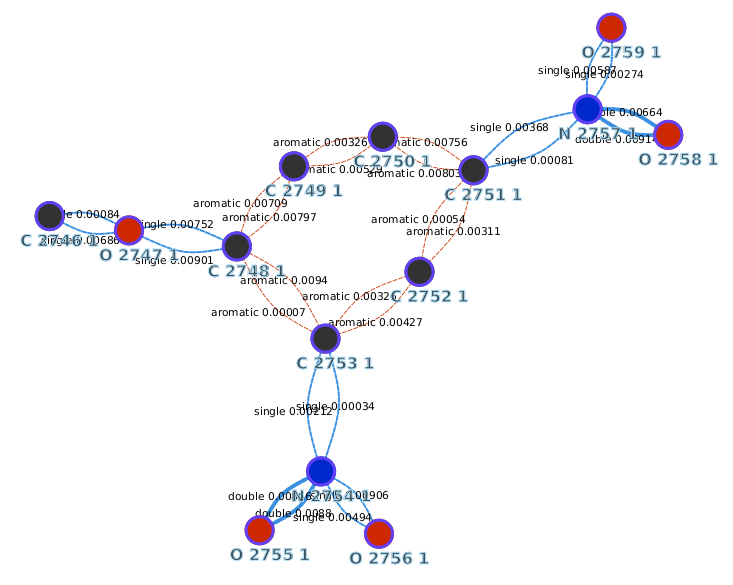}
    \caption{\footnotesize Example of graphs we consider in this paper where nodes and edges can contain rich information such as vector fields, bond types, importance, and much more.}
    \label{fig:mutag-sample}
\end{figure}

In this work, we introduce a novel, interpretable method for graph-level representation using sparse binary hypervectors and symbolic logic-based operations. Each node and edge is encoded as a high-dimensional sparse binary vector, where semantic properties (e.g., label, importance, and attribute values) are independently embedded. Information is propagated across the graph in a hierarchical fashion using symbolic binding: starting from a node, we bind its embedding to that of each outgoing edge, then further to the destination node, capturing a node $\rightarrow$ edge $\rightarrow$ neighbor path, as shown in Figure \ref{fig:mutag-sample}. Each such path contributes a distinct bound vector, and all such vectors are aggregated using a bundling operation into a fixed-size graph-level representation. This encoding pipeline can be extended across multiple layers of abstraction (e.g., substructures or motifs) and is amenable to transparent reasoning and local interpretation, aligning naturally with TM-based learning systems.

Our approach retains the structural richness of graph data while avoiding the need for gradient-based backpropagation or opaque deep network architectures. It is well-suited for applications that demand interpretability, symbolic composability, or learning on low-resource or privacy-sensitive data.

\subsection{Related Work}

The most closely related work to our approach is the recent and forthcoming \emph{Graph Tsetlin Machine} by Granmo et al.~\cite{granmo2024graphTM}, which proposes a symbolic learning framework for graphs using Tsetlin Automata and propositional logic. In their model, node and edge features are embedded as binary hypervectors, and logic-based clauses are learned to capture node-level patterns via message passing. The Graph TM operates over a fixed, global graph structure in which all nodes and edges are known in advance, and only the values (features or labels) associated with those entities vary across training samples. This makes the model particularly suitable for domains such as traffic networks, power grids, or knowledge graphs, where the topology is static but the observed signals evolve.

By contrast, our method addresses the more general and challenging problem of graph classification across a distribution of graphs with varying structure. Each sample in our setting is a standalone graph instance with potentially different nodes, edges, and topologies. To handle this, we propose a novel encoding strategy that maps each graph into a sparse binary hypervector through a symbolic hierarchy: nodes are bound to their label, importance, and attributes; edge types and attributes are bound and passed to destination nodes; and the resulting messages are bundled into a global graph-level representation. This approach enables TMs to operate in inductive settings where graphs differ not just in values but also in structure, extending symbolic reasoning to the full space of variable-topology graph classification tasks.

In essence, while both approaches leverage Tsetlin learning and symbolic hypervector encodings, they address fundamentally different problem settings: the Graph TM assumes a fixed graph and learns functions over evolving node/edge states, whereas our method encodes and learns over fully dynamic graph instances.

\subsection{Paper structure}
The remainder of this paper is organized as follows. In Section~\ref{sec:hypervectors}, we formally define sparse binary hypervectors and describe two key embedding mechanisms: linear scalar embeddings for continuous values, and interval-based symbolic embeddings for categorical variables. Section~\ref{sec:graphencoding} presents our graph encoding algorithm, which uses multi-hop symbolic binding to encode graph structure into a single fixed-length vector, and describes how these representations are used with the TM for graph classification. In Section~\ref{sec:interpretability}, we discuss the interpretability properties of our approach, including the ability to decompose global predictions into localized subgraph contributions. Section~\ref{sec:experiments} provides numerical comparisons on standard graph classification benchmarks from the TU Dortmund dataset collection~\cite{morris2020TUDataset}, where we evaluate our approach against established state-of-the-art models, including Graph Convolutional Networks (GCNs)~\cite{kipf2016semi} and Graph Attention Networks (GATs)~\cite{velickovic2018graph}. We conclude in Section~\ref{sec:conclusion} with a summary of our findings and suggestions for future extensions.

\section{Sparse Binary Hypervectors for Rich Graph Structures}
\label{sec:hypervectors}

Graphs often exhibit a rich variety of node- and edge-level characteristics. For example, nodes may carry symbolic labels (e.g., atomic species), scalar values such as centrality or importance scores, or multivariate continuous features. Similarly, edges may be annotated with categorical types (e.g., chemical bond types) or numerical attributes like distances or weights. Sparse binary hypervectors (SHV, see for example \cite{Kleyko_2022}) offer an attractive framework for encoding graph information due to their simplicity, robustness, and composability. Represented as high-dimensional binary vectors with only a small proportion of bits set to 1, they naturally support efficient storage and computation. Their binary structure makes them inherently noise-tolerant—small perturbations or missing features do not drastically affect similarity measures like Hamming distance. Furthermore, they enable a well-defined symbolic algebra: binding operations (e.g., XOR) preserve pairwise relationships, while bundling (e.g., majority vote) allows aggregation of multiple components (such as neighbor nodes or edge interactions) into a single, interpretable representation. This makes them especially suited for modeling complex graph structures while maintaining transparency and stability.

To support a uniform and compositional encoding of this heterogeneous information, we define embedding strategies that map all such features into sparse binary hypervectors. These embeddings ensure that diverse data types are represented in a consistent format, enabling symbolic binding and bundling operations to propagate structural and semantic information through the graph.

\subsection{Definition}

SHVs are defined as a high-dimensional binary vector $\mathbf{v} \in \{0,1\}^D$ with a fixed number of active (1-valued) bits. Let $K \ll D$ be a sparsity constant such that:
\[
\|\mathbf{v}\|_0 = K, \quad \text{where } \mathbf{v} \in \{0,1\}^D.
\]
These vectors will be used to represent symbols, attributes, or positions with robustness to noise and high capacity for compositional structure.

\subsection{Core Operations}

\paragraph{Binding ($\otimes$)}
The \emph{binding} operator $\otimes$ is used to combine two hypervectors into a new one representing their association. Binding is defined as element-wise XOR:
\[
\mathbf{v}_{\text{bind}} = \mathbf{v}_1 \otimes \mathbf{v}_2 := \mathbf{v}_1 \oplus \mathbf{v}_2.
\]
and is invertible, namely: $\mathbf{v}_1 \otimes \mathbf{v}_2 \otimes \mathbf{v}_2 = \mathbf{v}_1$.

\paragraph{Bundling ($\oplus$)}
The \emph{bundling} operator $\oplus$ aggregates a collection of hypervectors $\{\mathbf{v}_{1}, \dots, \mathbf{v}_{M}\}$ into a single prototype. For each bit position $j \in \{1, \ldots, D\}$, define:
\[
c_j = \sum_{i=1}^{M} v_{i,j}.
\]
We select the top-$K$ positions with the highest counts and define the bundled hypervector $\mathbf{v}_{\text{bundle}}$ as:
\[
v_{\text{bundle}, j} =
\begin{cases}
1 & \text{if } j \in \text{TopK}(c_1, \ldots, c_D) \\
0 & \text{otherwise}.
\end{cases}
\]
This preserves the most commonly active bits while maintaining sparsity.

We denote this operation compactly as:
\[
\mathbf{v}_{\text{bundle}} = \bigoplus_{i=1}^M \mathbf{v}_{i}.
\]

\subsection{Embedding Strategies}

To encode rich information on nodes and edges, we will use the following types of embeddings:

\subsubsection*{Linear Embedding}

The \textit{linear embedding} maps a real-valued scalar \( x \in [a, b] \subset \mathbb{R} \) into a sparse binary hypervector \( \mathbf{v}_x \in \{0,1\}^D \). The goal is to preserve continuity, namely nearby values should map to hypervectors with small Hamming distance (defined as the number of bits that differ bitwise between two vectors) and further away values should have larger Hamming distances.

We partition the interval \([a, b]\) into \( Q \) equally spaced subintervals:
\[
\Delta = \frac{b - a}{Q}, \quad \text{and} \quad x_i = a + i\Delta \quad \text{for } i = 0, \ldots, Q-1.
\]

A base hypervector \( \mathbf{v}_{x_0} \) is generated randomly:
\[
\mathbf{v}_{x_0} \sim \mathcal{H}(D, k),
\]
where \( \mathcal{H}(D, k) \) is the set of binary vectors of dimension \( D \) with exactly \( k \) ones.

Each subsequent vector \( \mathbf{v}_{x_{i+1}} \) is derived from \( \mathbf{v}_{x_i} \) by flipping a small proportion of bits to ensure gradual change:
\[
\mathbf{v}_{x_{i+1}} = \operatorname{FlipBits}(\mathbf{v}_{x_i}, \alpha D + \beta D),
\]
where:
- \( \alpha \in (0,1) \) is the continuity flip rate, used to flip \emph{previously set} bits for smooth transitions,
- \( \beta \in (0,1) \) is the noise rate, introducing additional randomness by flipping new, unrelated bits.

The total number of flipped bits per level is \( (\alpha + \beta)D \), typically small relative to \( D \), ensuring sparsity and local similarity:
\[
\text{Hamming}(\mathbf{v}_{x_i}, \mathbf{v}_{x_{i+1}}) \ll D.
\]

Given a scalar input \( x \in [a, b] \), we assign its embedding by locating the nearest partition:
\[
i = \left\lfloor \frac{x - a}{\Delta} \right\rfloor, \quad \text{and return } \mathbf{v}_x = \mathbf{v}_{x_i}.
\]

\paragraph{Interval and Categorical Embedding.}
To embed discrete values from a finite set $\{x_1, \ldots, x_Q\}$, each level $x_q$ is assigned a randomly generated sparse binary vector:
\[
\phi_{\text{interval}}(x_q) = \text{RandomSparseVector}(K).
\]
These vectors are nearly orthogonal, making them ideal for role embeddings or categorical features where similarity should not imply semantic proximity.

We will use the following embedding strategies for all the graph semantics in the rest of the paper:

\begin{table}[h!]
\centering
\caption{\footnotesize Embedding strategies for node and edge features in graph-structured data.}
\label{tab:embedding_strategies}
\begin{tabular}{@{}lll@{}}
\toprule
\textbf{Graph Feature} & \textbf{Example} & \textbf{Embedding Type} \\
\midrule
Node label             & Atom type, object class         & Categorical \\
Node attribute         & Power level, physical field     & Linear      \\
Node importance        & PageRank, centrality, position  & Interval    \\
Edge label             & Bond type, connection type      & Categorical  \\
Edge attribute         & Distance, cost, interaction strength & Linear \\
\bottomrule
\end{tabular}
\end{table}

\section{Graph Encoding via Sparse Hypervector Composition}
\label{sec:graphencoding}

We propose a graph encoding strategy that produces a fixed-length sparse binary hypervector for each graph, regardless of its structure or size. This strategy builds upon the symbolic compositionality of sparse hypervectors and operates in a message-passing fashion, where information flows outward from each vertex through its outgoing edges and is recursively bound and bundled into a final representation.

\subsection{Graph Representation and Features}

Let a graph be denoted by \( G = (V, E) \), where:
- \( V = \{v_1, \ldots, v_n\} \) is the set of nodes (or vertices, we use them interchangeably),
- \( E \subseteq V \times V \) is the set of directed edges.

Each node \( v \in V \) is associated with the following attributes:
\begin{itemize}
    \item A categorical label \( \ell_v \in \mathcal{L}_v \), encoded via a categorical sparse hypervector \( \phi_{\text{cat}}(\ell_v) \),
    \item A vector of scalar attributes \( \mathbf{a}_v \in \mathbb{R}^d \), embedded via linear embedding \( \phi_{\text{lin}}(\mathbf{a}_v) \),
    \item An importance score \( r_v \in [0,1] \), derived from a PageRank score and embedded via an interval embedding \( \phi_{\text{int}}(r_v) \).
\end{itemize}

Each edge \( e = (v, u) \in E \) is annotated with:
\begin{itemize}
    \item Symbolic label \( \ell_e \in \mathcal{L}_e \), encoded via a categorical embedding \( \phi_{\text{cat}}(\ell_e) \),
    \item Scalar or vector attribute \( a_e \in \mathbb{R} \), encoded (dimension-wise) via linear embedding \( \phi_{\text{lin}}(a_e) \),
    \item Fixed role vector \( \phi_{\text{embed}}(e) \in \{0,1\}^D \).
\end{itemize}

\subsection{Encoding Strategy}

Let \( \psi(v) \in \{0,1\}^D \) denote the encoded representation of node \( v \). We construct this using the following composition:
\[
\psi(v) := \phi_{\text{cat}}(\ell_v) \otimes \phi_{\text{lin}}(\mathbf{a}_v) \otimes \phi_{\text{int}}(r_v),
\]
where \( \otimes \) denotes the binding (XOR) operation defined in Section~\ref{sec:hypervectors}.

To encode the full graph, we simulate attention-style message passing from each vertex \( v \in V \). For each outgoing edge \( e = (v, u) \), we compute a triple binding of:
\begin{itemize}
\item Source vertex representation \( \psi(v) \),
\item Edge label and role vector \( \phi_{\text{cat}}(\ell_e) \otimes \phi_{\text{embed}}(e) \),
\item Target vertex representation \( \psi(u) \).
\end{itemize}

This message is defined as:
\[
\mathbf{m}_{v \rightarrow u} = \psi(v) \otimes \phi_{\text{cat}}(\ell_e) \otimes \phi_{\text{embed}}(e) \otimes \psi(u) \otimes \phi_{\text{embed}}(v),
\]
where the final binding with \( \phi_{\text{embed}}(v) \) ensures role specificity.

All such messages are aggregated using sparse bundling:
\[
\mathbf{v}_G = \bigoplus_{(v,u) \in E} \mathbf{m}_{v \rightarrow u},
\]
where \( \oplus \) is the sparse weighted bundling operator that retains the top \( K \) most frequent bits across messages, maintaining fixed sparsity.

\begin{figure}  
    \centering
    \includegraphics[width=3in]{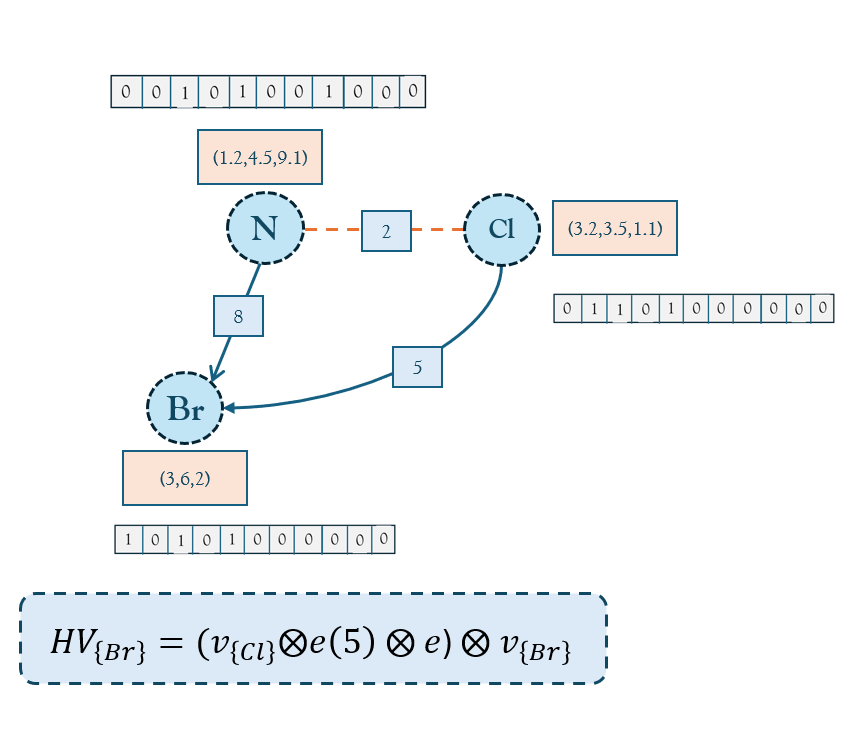}
    \caption{\footnotesize Example of encoding rich information from one node to parent node}
    \label{fig:encoding-nodes}
\end{figure}

Figure \ref{fig:encoding-nodes} illustrates the encoding mechanism from one node to a parent node. As each node label has a categorical role vector associated with it, this is used to bind with the attribute at the node, along with the edge information. Finally, to associate it with the target node, the role vector of the target node is binded to give locational information of the encoding.

\subsection{Ordering and PageRank Importance}

To prioritize informative message pathways, the nodes are sorted in descending order according to their importance scores \( r_v \), as estimated by the centrality of PageRank~\cite{page1999pagerank}. This ensures that information from structurally central nodes contributes earlier and more directly to the bundled graph-level representation.

The final graph hypervector \( \mathbf{v}_G \) serves as a compact symbolic representation of the full graph \( G \), encoding its topology and attribute semantics in a fixed-length, sparse binary form suitable for downstream classification using TMs.

\subsection{Layered Composition and Recursive Depth}

The encoding strategy we adopt is naturally hierarchical and can be organized into discrete layers of abstraction. In this work, we implement a two-layered message-passing architecture:

\begin{itemize}
    \item \textbf{Layer 1 (Node Encoding):} Each node \( v \in V \) is independently encoded as a sparse binary hypervector:
    \[
    \psi(v) = \phi_{\text{cat}}(\ell_v) \otimes \phi_{\text{lin}}(\mathbf{a}_v) \otimes \phi_{\text{int}}(r_v).
    \]
    \item \textbf{Layer 2 (Edge Composition with Parent):} For each outgoing edge \( e = (v,u) \in E \), a message is computed by binding the source node \( v \), the edge metadata, and the target node \( u \), resulting in:
    \[
    \mathbf{m}_{v \rightarrow u} = \psi(v) \otimes \phi_{\text{cat}}(\ell_e) \otimes \phi_{\text{embed}}(e) \otimes \psi(u) \otimes \phi_{\text{embed}}(v).
    \]
    All such messages are bundled into the final graph-level representation.
\end{itemize}

This layered construction enables relational structure to be encoded symbolically while respecting node roles and edge directions.

\paragraph{Extending to Three Layers.} A natural extension of this approach is to recursively propagate the encoding to a third layer by considering the neighbors of the target node \( u \). In such a model:
\begin{itemize}
    \item Each node’s Layer 2 encoding \( \mathbf{m}_{v \rightarrow u} \) is treated as the input to the next iteration.
    \item The message from \( u \) to a subsequent neighbor \( w \) incorporates \( \mathbf{m}_{v \rightarrow u} \), the edge \( e' = (u,w) \), and the hypervector for \( w \):
    \[
    \mathbf{m}_{u \rightarrow w} = \mathbf{m}_{v \rightarrow u} \otimes \phi_{\text{cat}}(\ell_{e'}) \otimes \phi_{\text{embed}}(e') \otimes \psi(w).
    \]
\end{itemize}

This results in a three-layered graph representation that captures longer-range dependencies and nested relational structures. While this introduces additional computational cost, it provides deeper context propagation and may be beneficial in large graphs with low local connectivity or high semantic drift.

The number of layers in this framework can be viewed analogously to the receptive field in convolutional neural networks, where additional layers allow information to diffuse across increasingly distant regions of the graph.

\section{Graph Classification with Tsetlin Machines}
\label{sec:classification}

To classify these graph-level representations, we employ the standard Coalesced TM (CoTM) \cite{glimsdal2021coalescedmultioutputtsetlinmachines}. This architecture allows for efficient representation of the clauses while building weights to derive class sums.

After each input graph \( G \) is encoded as a sparse binary hypervector \( \mathbf{v}_G \in \{0,1\}^D \) constructed using the binding and bundling strategy detailed in Section~\ref{sec:graphencoding}, this vector serves as the input to the TM, where each bit position \( \mathbf{v}_G[i] \in \{0,1\} \) is treated as a literal. Let \( \mathcal{C} = \{C_1, C_2, \ldots, C_T\} \) denote the shared set of clauses, where each clause \( C_t \) is a conjunction over a subset of literals and their negations:
\[
C_t(\mathbf{v}_G) = \bigwedge_{i \in \mathcal{I}_t^+} \mathbf{v}_G[i] \,\wedge \bigwedge_{j \in \mathcal{I}_t^-} \neg \mathbf{v}_G[j],
\]
where \( \mathcal{I}_t^+ \subseteq [1,D] \) and \( \mathcal{I}_t^- \subseteq [1,D] \) are index sets for included positive and negative literals respectively.

Each clause \( C_t \) is shared across all output classes \( y \in \{1, \ldots, K\} \), and is associated with a learned integer weight \( w_{t,y} \in \mathbb{Z} \). The classification function for class \( y \) computes a sum over all clause evaluations weighted by their learned contributions:
\[
f_y(\mathbf{v}_G) = \sum_{t=1}^{T} w_{t,y} \cdot C_t(\mathbf{v}_G).
\]

The predicted class \( \hat{y} \) is obtained by selecting the class with the largest aggregated clause response:
\[
\hat{y} = \arg\max_{y \in \{1, \ldots, K\}} f_y(\mathbf{v}_G).
\]

\section{Interpretability and Explanation}
\label{sec:interpretability}

One of the key advantages of the CoTM is the potential for intrinsic interpretability \cite{blakely2020closedformexpressionsgloballocal}. In contrast to opaque neural graph architectures, our hypervector-based approach allows for direct attribution of predictions to specific symbolic structures in the input graph.

\subsection{Local Interpretability via Clause Deconstruction}

Given a predicted class \( \hat{y} \) for a graph \( G \), the prediction is obtained via:
\[
\hat{y} = \arg\max_{y \in \{1,\dots,K\}} \sum_{t=1}^{T} w_{t,y} \cdot C_t(\mathbf{v}_G),
\]
where each clause \( C_t \) evaluates a subset of bits from the sparse binary graph hypervector \( \mathbf{v}_G \).

To interpret the prediction, we seek to answer:
\begin{center}
\emph{Which symbolic components (nodes, edges, attributes) most strongly contributed to this decision?}
\end{center}

\subsection{Backward Decoding: Identifying Influential Nodes}

Recall from Section~\ref{sec:graphencoding} that \( \mathbf{v}_G \) is produced by bundling a set of bound node-edge messages:
\[
\mathbf{v}_G = \bigoplus_{v \in V} \bigoplus_{(v,u) \in E} \mathbf{m}_{v \rightarrow u}.
\]

Each message \( \mathbf{m}_{v \rightarrow u} \) is composed as:
\[
\mathbf{m}_{v \rightarrow u} = \psi(v) \otimes \phi(\ell_e) \otimes \phi(e) \otimes \psi(u) \otimes \phi(v),
\]
with \( \psi(v) \) being the encoding of the source node and \( \phi(v) \) its role vector (importance embedding).

To identify the most influential node toward a specific prediction, we:
\begin{enumerate}
    \item Deconstruct the prediction by isolating the set of activated clauses \( \{ C_t \} \) with non-zero weights \( w_{t,\hat{y}} \).
    \item Use these clauses to form a partial reconstruction \( \mathbf{v}_{\text{pred}} \subseteq \mathbf{v}_G \), representing bits most used in the prediction.
    \item Iterate over all nodes \( v \in V \) and compute candidate messages \( \mathbf{m}_v \) by rebinding \( \phi(v) \) (the importance hypervector) with potential node-edge encodings from the graph.
    \item Compute the Hamming distance between \( \mathbf{m}_v \) and \( \mathbf{v}_{\text{pred}} \). The node with the minimal distance is inferred to have the strongest causal influence on the classification:
    \[
    v^* = \arg\min_{v \in V} \operatorname{Hamming}(\mathbf{m}_v, \mathbf{v}_{\text{pred}}).
    \]
\end{enumerate}

\subsection{Attribution through Role-Specific Decoding}

The node importance hypervector \( \phi(v) \), derived from interval embeddings (e.g., PageRank rank), serves as a positionally unique identifier in the hypervector space. This allows the model to reconstruct not just what pattern contributed to the output, but which node in the graph most likely originated it.

This interpretability pipeline enables symbolic tracing of classification logic, turning predictions into human-auditable rule chains over nodes, edges, and their attributes — a capability not typically present in neural graph models.

\subsection{Example: Decoding the Most Influential Node}

To demonstrate how our symbolic interpretability strategy operates in practice, we walk through a simplified example with a small hypervector space.

\paragraph{Setup.} Assume the dimensionality of the sparse binary hypervector is \( D = 12 \), and the sparsity level is set to \( s = 33\% \), meaning each bundled hypervector should contain exactly 4 active bits.

Suppose that for a given graph \( G \), the TM outputs class \( \hat{y} \) by activating a subset of clauses. The union of literals contributing to this prediction can be aggregated into a vector \( \mathbf{v}_{\text{pred}}^{\text{sum}} \in \mathbb{N}^{12} \), representing how frequently each bit was used across the active clauses:
\[
\mathbf{v}_{\text{pred}}^{\text{sum}} = [10,\, 20,\, 50,\, 80,\, 23,\, 1,\, 0,\, 0,\, 0,\, 1,\, 45,\, 0].
\]

To obtain a binary interpretation of the most relevant bits, we keep the top \( s \times D = 4 \) entries. This yields the index set of the 4 most voted bits:
\[
\text{Top indices} = \{3,\, 2,\, 10,\, 4\}.
\]
and thus the explanation vector becomes:
\[
\mathbf{v}_{\text{pred}} = [0,\,0,\,1,\,1,\,1,\,0,\,0,\,0,\,0,\,0,\,1,\,0].
\]
Suppose the graph has three nodes with known importance hypervectors \( \phi(v_i) \) as follows:
\[
\begin{aligned}
\phi(v_1) &= [0,\,0,\,1,\,0,\,1,\,1,\,0,\,0,\,0,\,0,\,1,\,0], \\
\phi(v_2) &= [0,\,0,\,0,\,0,\,0,\,0,\,0,\,0,\,1,\,1,\,1,\,1], \\
\phi(v_3) &= [0,\,0,\,1,\,1,\,1,\,0,\,0,\,0,\,0,\,0,\,1,\,0]. \\
\end{aligned}
\]
We compute the Hamming distance between each node's importance vector and the explanation vector:
\[
\begin{aligned}
\operatorname{Hamming}(\mathbf{v}_{\text{pred}}, \phi(v_1)) &= 3, \\
\operatorname{Hamming}(\mathbf{v}_{\text{pred}}, \phi(v_2)) &= 6, \\
\operatorname{Hamming}(\mathbf{v}_{\text{pred}}, \phi(v_3)) &= \mathbf{0}.
\end{aligned}
\]

Hence, node \( v_3 \) is determined to be the most influential component in the model's prediction. This node's encoded message contributed most strongly to the active clauses determining \( \hat{y} \), and its influence was preserved and traceable via symbolic binding with its role vector \( \phi(v_3) \).

\begin{figure*}[t]  
    \centering
    \includegraphics[width=\textwidth]{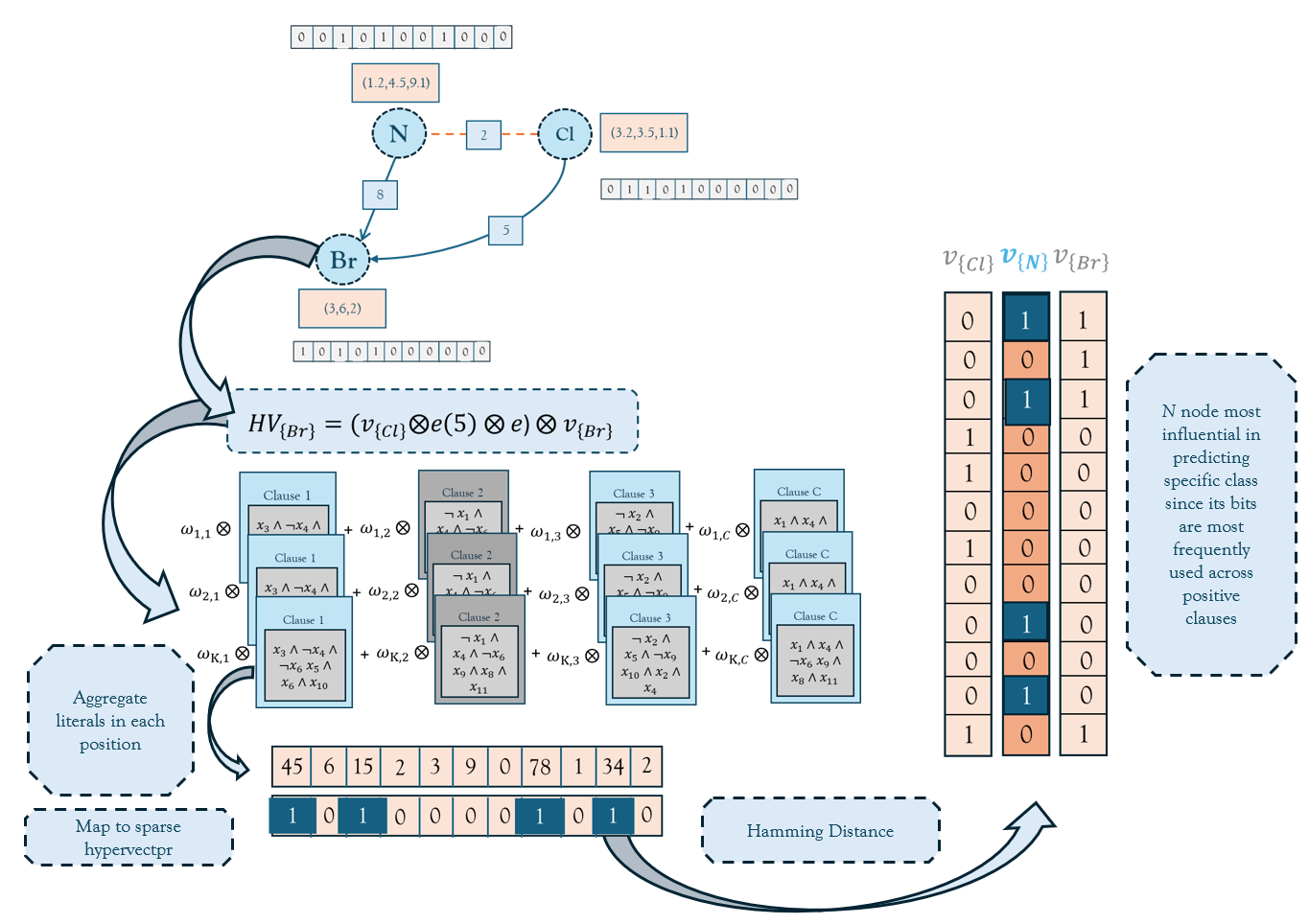}
    \caption{\footnotesize The process of predicting graph class and then decoding which node provided the most impact using Hamming distance.}
    \label{fig:mutag-interpretability}
\end{figure*}

Figure \ref{fig:mutag-interpretability} illustrates the process of predicting graph class and then decoding which node provided the most impact using Hamming distance. The clauses which had an output of 1 are shown in blue and the aggregated across all these clauses resulting in a count of each literal from the hypervector space. We then transform this to a hypervector from our SHV space. We then compare this to all attributes in our graph encoding scheme. For simplicity, we consider only node label and importance encodings.

\section{Numerical Results}\label{sec:experiments}

\subsection{Benchmark Datasets from TUDataset}

To evaluate our multilayered symbolic graph learning framework, we conduct experiments on benchmark datasets from the \textbf{TUDataset} collection~\cite{morris2020TUDataset}, a standard suite for graph classification. These datasets are widely used for evaluating both kernel-based and neural graph models. Each graph sample may differ in topology, number of nodes and edges, as well as in the presence of symbolic labels or numerical attributes on nodes and edges.

\begin{figure}[ht]
    \centering
    \begin{subfigure}[b]{0.45\textwidth}
        \centering
        \includegraphics[width=\textwidth]{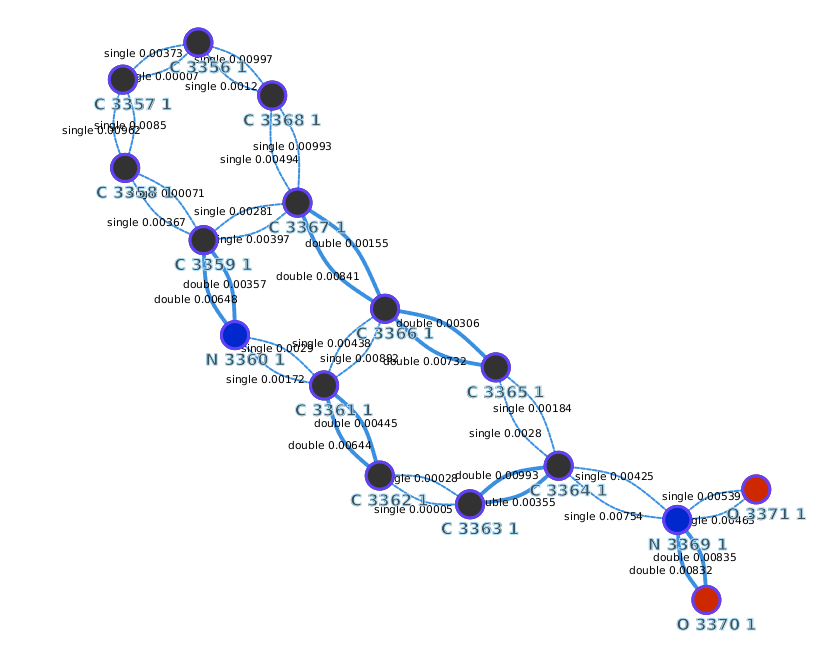}
        \caption{\footnotesize Sample graph from the negative class (non-mutagenic compound).}
        \label{fig:mutag-negative}
    \end{subfigure}
    \hfill
    \begin{subfigure}[b]{0.45\textwidth}
        \centering
        \includegraphics[width=\textwidth]{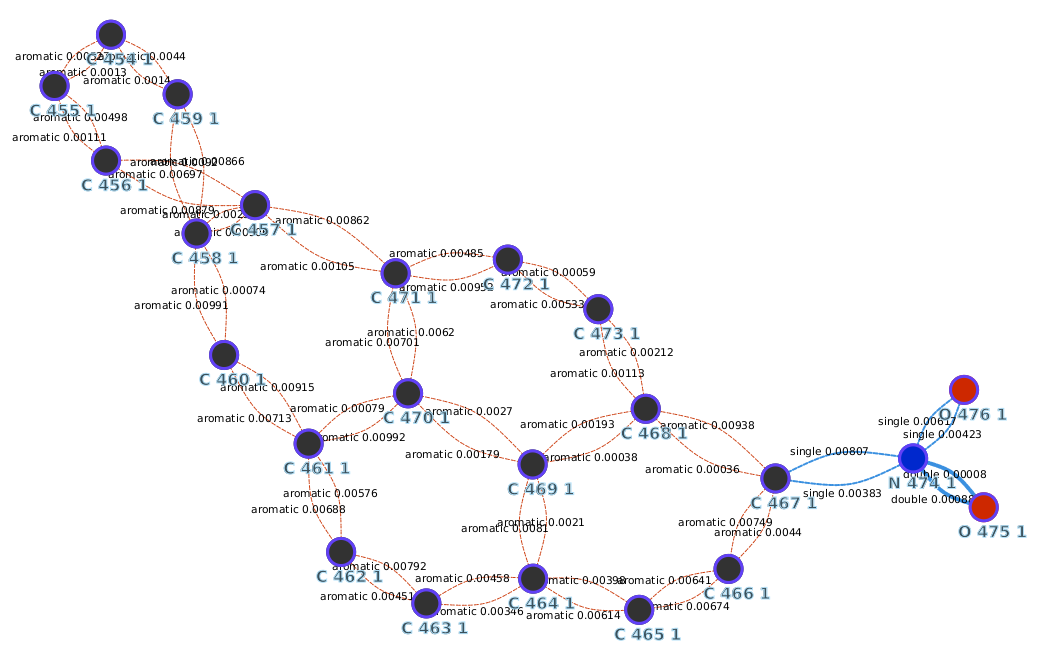}
        \caption{\footnotesize Sample graph from the positive class (mutagenic compound).}
        \label{fig:mutag-positive}
    \end{subfigure}
    \caption{\footnotesize Visualizations of two sample graphs from the MUTAG dataset. Nodes are colored according to their atomic type, and edges their bond type.}
    \label{fig:mutag-samples}
\end{figure}

We focus on seven representative datasets: AIDS, DHFR, DHFR\_MD, and NCI1, MUTAG, ER\_MD, and PROTEINS which offer varying levels of complexity in symbolic structure, node/edge features, and graph scale. Figure \ref{fig:mutag-samples} shows two examples from the MUTAG class, with one representing a non-mutagenic compound and the other a mutagenic compound. 

All experiments were conducted using a fixed hypervector dimensionality and TM configuration to ensure consistent evaluation across datasets. We use sparse binary hypervectors of dimension $D = 6400$, represented internally as arrays of 100 long integers (each storing 64 bits). The hypervectors are initialized with a fixed sparsity level of 20\%, ensuring each vector contains approximately 1280 active bits. Our CoTM model uses a fixed set of hyperparameters for all experiments
\begin{itemize}
    \item Clauses: 500
    \item Threshold ($T$): 1000
    \item Specificity ($s$): 2.0
    \item Max Literals per Clause: 50
\end{itemize}

Each experiment is run for 4 training epochs. For evaluation, we randomly split each dataset into 70\% training and 30\% testing subsets. To account for randomness in training and data splits, we repeat each experiment 10 times with different random seeds and report the mean and standard deviation of the test accuracies.

No hyperparameter optimization was performed. All parameters were held constant across datasets to emphasize the generality and robustness of our symbolic graph encoding and classification approach.

\begin{itemize}
\item The AIDS dataset consists of molecular graphs with nodes representing atoms and edges representing chemical bonds. Each node has a discrete label corresponding to atom type with no continuous node or edge attributes. Accurate classification requires distinguishing graphs based on symbolic node and bond-type patterns, despite their structural similarities.

\item The DHFR dataset contains molecular graphs from dihydrofolate reductase compounds. Graphs include atom-type node labels and bond-type edge labels.The model must learn symbolic bond structure patterns from moderately sized graphs with diverse molecular arrangements.

\item The DHFR\_MD dataset enhances DHFR by adding 3D molecular structure. Each graph is a complete graph over atoms, with edge attributes encoding Euclidean distances, and edge labels describing bond types (e.g., single, double, aromatic). The symbolic model must integrate both discrete and continuous edge information into a unified encoding, increasing the difficulty of capturing relevant structural motifs.

\item The NCI1 dataset is a large-scale chemical compound dataset for binary classification. Each graph contains atom-type node labels and bond-type edge labels. The NCI1 dataset is a large-scale chemical compound dataset for binary classification. Each graph contains atom-type node labels and bond-type edge labels, but no other attribute information.

\item The MUTAG dataset describes mutagenic aromatic and heteroaromatic nitro compounds. It contains small molecular graphs typically labeled with mutagenicity outcomes.

\item ER\_MD is a molecular dataset similar in spirit to the *-MD revisited collections, containing full molecular graphs annotated with 3D distances as numeric edge attributes, and standard atom/bond labels.

\item PROTEINS is a set of graphs that correspond to a whole protein, with the goal being to classify the protein into one of two categories enzyme vs. non-enzyme. Node labels are helix and sheet encoded as categorical labels and has a feature vector (numeric attributes) that captures structural and physicochemical properties such as the length of the SSE, hydrophobicity, and other bio-relevant characteristics.

\end{itemize}

\begin{table}[h!]
\centering
\caption{\footnotesize Summary of graph dataset features used in our experiments.}
\label{tab:dataset-features}
\begin{tabular}{lcccc}
\toprule
\textbf{Dataset} & \textbf{Node Labels} & \textbf{NodeAttr} & \textbf{Edge Labels} & \textbf{EdgeAttr} \\
\midrule
AIDS      & atom type      & Chem/X/Y                        & bond          & No \\
DHFR      & atom type      & X/Y                        & No         & No \\
DHFR\_MD  & atom type      & No                        & bond          & distance \\
NCI1      & atom type      & No                        & No                      & No \\
ER\_MD     & atom type      & No                        & bond                      & distance \\
MUTAG     & atom type      & No                        & bond                      & No \\
PROTEINS & SSE & chemical & No  & No \\
\bottomrule
\end{tabular}
\end{table}

\begin{table}[h!]
\centering
\caption{\footnotesize Graph classification accuracy (\%) on benchmark datasets. Bold on SGI-TM represents competitive results with other published benchmarks}
\label{tab:comparison}
\begin{tabular}{lcccc}
\toprule
\textbf{Dataset} & \textbf{SGI-TM (Ours)} & \textbf{Graph Kernel} & \textbf{GNN}  \\
\midrule
AIDS       & 71.2~$\pm$~3.4 & WL: $\approx$84~\cite{Shervashidze2011WL} & GCN: $\approx$74.0~\cite{xu2019powerful} \\
DHFR\_MD   & \textbf{89.0~$\pm$~2.2} & SP: $\approx$73.5~\cite{Bai2015MWSP} & GNN: $\approx$88~\cite{Zhu2020IHGNN}  \\
DHFR       & 79.0~$\pm$~4.0 & WL: $\approx$84~\cite{Shervashidze2011WL} & GIN: $\approx$79~\cite{xu2019powerful}  \\
NCI1       & 61.3~$\pm$~2.0 & WL: $\approx$86~\cite{Zhang2018DGCNN} & GIN:  $\approx$82.7~\cite{xu2019powerful}  \\
MUTAG     & 90.1~$\pm$~3.6 & WL: $\approx$84~\cite{Shervashidze2011WL}              & GNN: $\approx$92.6~\cite{xu2019powerful}  \\
ER\_MD    & \textbf{84.0~$\pm$~5.1} & SM:$\approx$82.8~\cite{kriege2012subgraphmatchingkernelsattributed}  & GPN:$\approx$78.26 \cite{tang2020adversarialattackhierarchicalgraph}             \\
PROTEINS   & \textbf{77.2~$\pm$~1.8}   & $\approx$76.2~\cite{neumann2014propagationkernels} & GIN:76.0$\pm$3.2~\cite{xu2019powerful} \\
\bottomrule
\end{tabular}
\end{table}

\subsection{Summary of Results}

Across all datasets tested, our SGI-TM demonstrates competitive performance, particularly on datasets with richer structural features. In particular, the DHFR\_MD, ER\_MD, and PROTEINS datasets yielded the highest classification accuracies among our experiments, achieving results comparable to state-of-the-art kernel methods and graph neural networks (GNNs).

A common trait among these datasets is the presence of node or edge attributes, in addition to symbolic labels. We hypothesize that the inclusion of continuous or high-dimensional node/edge attributes provides greater representational diversity, enabling the clause-based learning in TMs to identify finer-grained logical patterns. Since our symbolic hypervector embeddings incorporate both categorical and numerical information in a uniform high-dimensional binary space, they are particularly well-suited for learning in heterogeneous and richly annotated graph structures. This may explain the superior performance of SGI-TM on these datasets relative to those with only discrete labels and no additional attributes, such as the NCl1 dataset which performed by far the worst. 

These results suggest that symbolic clause-based learning may offer a robust and interpretable alternative to deep neural graph architectures, especially when structured attribute information is available.

\section{Conclusion}\label{sec:conclusion}

In this work, we introduced a symbolic and interpretable approach to graph classification using sparse binary hypervectors and TMs. Our method encodes rich structural information from graphs—including symbolic node and edge labels, continuous-valued attributes, and node importance—into a uniform hyperdimensional representation. These representations are then processed by a coalesced multi-output TM, offering clause-based learning that is both symbolic and interpretable.

Compared to state-of-the-art kernel and GNN approaches, our method exhibits several notable strengths. It is fast to train, requires no backpropagation, and is highly parallelizable due to the decentralized nature of TM learning. Most importantly, it offers clear local interpretability by tracing clause activation back to contributing node-edge structures, which is typically opaque in neural methods. Additionally, our symbolic hypervector encoding allows for seamless integration of categorical, ordinal, and continuous features within a unified algebraic framework.

However, the approach is not without limitations. Hyperparameter of TM selection remains challenging, and in this study we opted to use a fixed configuration across all datasets to avoid overfitting and reduce tuning bias. While the model demonstrates competitive performance—particularly when graphs include rich attribute information—it does not fully exploit message-passing mechanisms that have proven effective in GNN architectures. Adapting TMs to support more expressive message-passing for general graph classification remains an open question.

For future work, we envision extending this framework by identifying and encoding reusable subgraph structures (motifs), and learning clause patterns over these subcomponents. This may allow us to combine symbolic interpretability with localized structural reuse, thereby improving classification accuracy while retaining transparency. Another avenue is clause specialization, where different clause families focus on different structural roles (e.g., hubs, bridges, motifs), allowing the machine to reason over heterogeneous graph topologies more effectively.

\bibliographystyle{unsrt}  
\bibliography{references}

\end{document}